\documentclass[10pt]{article}

\usepackage[a4paper,textwidth=18cm,textheight=24cm,top=2.85cm, bottom=2.85cm, left=1.5cm, right=1.5cm]{geometry}

\usepackage{icdp2009}

\makeatletter
\long\def\@makecaption#1#2{%
\vskip\abovecaptionskip
\sbox\@tempboxa{#1. #2}%
\ifdim \wd\@tempboxa >\hsize
#1. #2\par
\else
\global \@minipagefalse
\hb@xt@\hsize{\box\@tempboxa\hfil}%
\fi
\vskip\belowcaptionskip}
\makeatother
\usepackage[misc]{ifsym}
\usepackage{lmodern}
\usepackage{amsmath}
\usepackage{graphicx}
\usepackage{times}
\usepackage{subcaption}
\usepackage{float}
\usepackage{booktabs}
\usepackage{multirow}
\begin{document}
\noindent

% This should produce references in the order they appear
% \bibliographystyle{ieeetr}
\title{Unsupervised Classification of Street Architectures \\ Based on InfoGAN}

\author{\large \textbf{Ning Wang\textsuperscript{1}, Xianhan Zeng\textsuperscript{1}, 
Renjie Xie\textsuperscript{1},
Zefei Gao\textsuperscript{1}, Yi Zheng\textsuperscript{2}, Ziran Liao\textsuperscript{2},\\
Junyan Yang\textsuperscript{2} and Qiao Wang\textsuperscript{1}}\Letter \\[5mm]
\large \textsuperscript{1}School of Information Science and
Engineering and Shing-Tung Yau Center, \\Southeast University, Nanjing, 210096, China\\
\{wang\_ning, xianhan\_zeng, renjie\_xie, zf\_gao, qiaowang\}@seu.edu.cn\\
\textsuperscript{2}School of Architecture, Southeast University, Nanjing, 210096, China\\
zhengy17\_seu@163.com, natureliao1101@gmail.com, yangjy\_seu@163.com}

% \authoraddr{\textsuperscript{1}School of Information Science and
% Engineering and Shing-Tung Yau Center, \\Southeast University, Nanjing, 210096, China\\
% \{wang\_ning, xianhan\_zeng, renjie\_xie, zf\_gao, qiaowang\}@seu.edu.cn\\
% \textsuperscript{2}School of Architecture, Southeast University, Nanjing, 210096, China\\
% zhengy17\_seu@163.com, natureliao1101@gmail.com, yangjy\_seu@163.com
% }
\maketitle
%optional 
% \secauthoraddr{\textsuperscript{2}School of Architecture, Southeast University, Nanjing, 210096, China\\
% zhengy17\_seu@163.com, natureliao1101@gmail.com, yangjy\_seu@163.com}

\keywords
%Unsupervised classification, InfoGAN, FCAN, Street architectures, Streetscape analysing.
Unsupervised Classification, Information Maximizing GAN, Streetscape Analysing
\abstract
Street architectures play an essential role in city image and streetscape analysing.  However, existing approaches are all supervised which require costly labeled data. To solve this, we propose a street architectural unsupervised classification framework based on Information maximizing Generative Adversarial Nets (InfoGAN), in which we utilize the auxiliary distribution $Q$ of InfoGAN as an unsupervised classifier. Experiments on database of true street view images in Nanjing, China validate the practicality and accuracy of our framework. Furthermore, we draw a series of heuristic conclusions from the intrinsic information hidden in true images. These conclusions will assist planners to know the architectural categories better.

% Furthermore, from the intrinsic information hidden in true images we draw a series of heuristic conclusions, which will assist planner to know the architectural categories better.

\section{Introduction}
%, Green View Index (GVI)\cite{clay2004assessing},Visual Entropy (VE)\cite{ozkan2014assessment}
Streetscapes represent the most direct connection between urban residents and the city image, since they are the most prolific public spaces in the urban landscape and are an elemental setting for everyday activity[1]. Since Kevin Lynch first proposed the concept of City Image[2], many indicators have been put forward to describe the physical attributes of streetscapes, such as Salient Region Saturation (SRS)[3], and Sky-Openness Index (SOI)[4], etc. These indicators do contribute to evaluating the visual perception of streetscapes to some extent, however it is not adequate since they do not consider the significant impact of architectures in the streetscape.

Diverse styles and genres of urban buildings play an essential role in describing the global visual perception of streetscapes. However, it takes challenges to correctly and objectively classify architectural categories since they are determined by a large number of overlapping factors neither quantified nor qualified. Fortunately the recent rapidly developing technologies in machine learning provide new approaches, consist of two main paradigms, say, the supervised classification and unsupervised clustering. Both of them are based on street view images data, which contains a large quantity of information on urban streetscapes and is commonly convenient to gain.
%\cite{yin_big_2015}.

Supervised classification methods are based on manually labeled data with high acquisiting cost. Manual labeling is the common approach to gain training data. In this way, we first crawl unlabeled street view images from street view services such as Google[5], etc, and then these images are labeled manually. Note that the prior knowledge referred to in manually labelling are inevitable incorrect due to subjective factors. That is to say, the databases used for supervised classification have two disadvantages: the expensive cost, as well as the ambiguity of human subjectivity.

Because of the above two defects, we turn attention to unsupervised clustering researches. The fact that intelligent creatures, like humans, are able to classify things without supervision makes it plausible to assume that unsupervised methods can be developed to extract the intrinsic information on classification. Thus in this way hidden architectural category features are able to be mined from data, instead of defined by non-uniform prior knowledge. 

The contributions of this paper lie in three aspects:\\
\\
% \item{a. We propose a street architectural clustering framework by combining and improving existing technologies such as Fully Convolutional Adaptation Networks (FCAN) and InfoGAN. Experiments have indicated that our framework is practical to cluster street architectures correctly objectively.}
% \item{b. Through our framework the intrinsic categorical information of street architectures is mined, from which many heuristic conclusions are drawn and this knowledge will provide a new perspective to understand the interaction between street buildings and human perceptions.}
% \item{c. We implement the InfoGAN in fields of streetscape measuring and analysing for the first time. That is to say, the framework proposed in this paper can inspire potential applications of InfoGAN in unsupervised classification tasks. }\\
a. We propose a street architectural clust
ering framework by combining and improving existing technologies such as Fully Convolutional Adaptation Networks (FCAN) and InfoGAN. Experiments have indicated that our framework is practical to cluster street architectures correctly objectively.\\
b. Through our framework the intrinsic categorical information of street architectures is mined, from which many heuristic conclusions are drawn and this knowledge will provide a new perspective to understand the interaction between street buildings and human perceptions.\\
c. We implement the InfoGAN in fields of streetscape measuring and analysing for the first time. That is to say, the framework proposed in this paper can inspire potential applications of InfoGAN in unsupervised classification tasks. \\

In section 2 a brief overview of the embedding techniques as well as the used neural network classifiers are introduced. In section 3 we describe the methods we use in our framework. In section 4 we present some experiments and analyse the result. This work concludes in section 5 with a discussion of directions for future works.

\section{Relative works}
\subsection{Semantic Segmentation}
%DeepLab\cite{chen_deeplab:_2016}
As one of the most challenging tasks in computer vision, semantic segmentation aims to predict pixel-level semantic labels of the given image or video frame. In the context of the recent advance of Fully Convolution Networks (FCN)[6], several techniques,  ranging from multi-scale feature ensemble (e.g., Dilated Convolution, RefineNet, and HAZNet[7]) to context information preservation (e.g., ParseNet[8], PSPNet[9], DeepLab[10] and DST-FCN[11]), have been proposed. Meanwhile the original FCN formulation could also be improved by exploiting some post processing techniques (e.g., conditional random fields). Furthermore, recent work has tried to leverage weak supervision such as instance-level bounding boxes and image-level tags for semantic segmentation task, instead of relying on the pixel-level annotations which require expensive labeling efforts. 

%On the other hand, the Deep Domain adaptation aims to transfer model learnt in a labeled source domain to a target domain in a deep learning framework.The research of this topic has proceeded along three different dimensions: unsupervised adaptation, supervised adaptation and semi-supervised adaptation. Unsupervised domain adaptation refers to the setting when the labeled target data is not available. Deep Correlation Alignment (CORAL)\cite{sun_return_2015} exploits Maximum Mean Discrepancy (MMD) to match the mean and covariance of source and target distributions. Adversarial Discriminative Domain Adaptation (ADDA)\cite{tzeng_adversarial_2017} optimizes the adaptation model with adversarial training.

In this paper we mainly focuses on unsupervised adaptation for semantic segmentation task. Although this topic is seldom investigated, there are some works closely related to our theme. The method we utilize in our work mainly refers to the Fully Convolutional Adaptation Networks (FCAN) in [12] and FCNWild in [13], which addresses domain adaptation problem from the viewpoint of both visual appearance-level and representation-level domain invariance and is potentially more effective at undoing the effects of domain shift.

\subsection{Unsupervised Calssification}
%(according to \cite{springenberg_unsupervised_2015}), 
Unsupervised classification has traditionally been formalized as a clustering problem, for which some classical methods (e.g., K-means, Gaussian mixture, and density estimation) were developed. Later, several discriminative approaches  were explored. They transform the data into some form of low-dimensional representation before clustering. Nevertheless, these approaches easily overfit spurious correlations in the training data and that misguides classification decisions.

Meanwhile, a significant fraction of unsupervised learning research is driven by Generative Modelling. It is motivated by the belief that the ability to synthesize, or “create” the observed data entails some form of understanding, and it is hoped that a good generative model will automatically learn a disentangled representation, even though it is easy to construct perfect generative models with arbitrarily bad representations. Early models include Deep Boltzmann Machines[14], Stochastic generative networks and Autoencoders[15]. They circumvent the overfitting problem since they do not need to fit at all, thereby making learned representation more reliable. The most prominent generative models are the Variational Autoencoder (VAE)[16] and the Generative Adversarial Network (GAN)[17]. Although researches on generative models themselves are heated recently, their applications on unsupervised classification has not been thoroughly explored.
 
\section{Method}
In this section, we introduce in detail our proposed framework for street architectures unsupervised classification. This is the first time InfoGAN is applied in fields of streetscape measuring and analysing.
\subsection{FCAN}
\begin{figure}
    \centering
    \includegraphics[width=8cm]{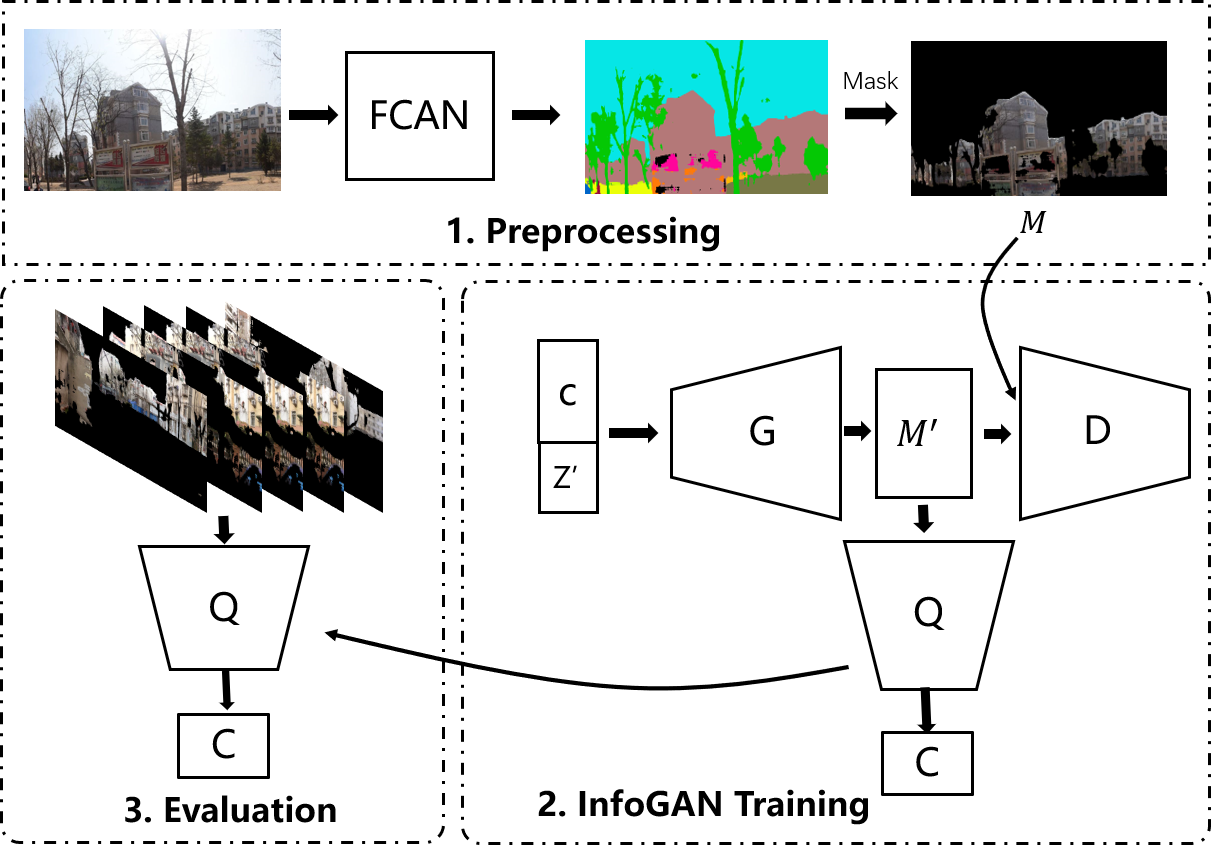}
    \caption{Street Architectural Classification Framework}
    \label{fig:3-1}
\end{figure}
The approach we use in the process of semantic segmentation is mainly inspired by the FCAN in [12]. It consists of two components: Appearance Adaptation Networks (AAN) and Representation Adaptation Networks (RAN). AAN is first utilized to transfer images from one domain to the other from the perspective of visual appearance. By recombining the image content in one domain with the “style” from the other one, the visual appearance tends to be domain-invariant. On the other hand, RAN learns domain-invariant representations in an adversarial manner and a domain discriminator is devised to classify which domain the image region corresponding to the receptive field of each spatial unit in the feature map comes from. As a result, the accuracy of semantic segmentation of street view images is much higher than it achieved by FCN and other methods.

\subsection{InfoGAN}

%Since Goodfellow \textit{et al}. proposed Generative Adversarial Network(GAN) in 2014\cite{goodfellow_generative_nodate}, a large number of derived models of GAN with better performance have been presented successively. Info-GAN is one of them.

%However, when comes to high-dimensional data such as images, the computational complexity becomes daunting and the performance deteriorates rapidly.
%When comes to images, the index is some statistic of distances between pixels of images. However, what we attempt to gain are the connections between the deep features of street view images, which are not able be reflected by distance.
Common approaches for unsupervised clustering tasks are based on distances, which are used as an index to evaluate the similarity between samples of train data. However, when comes to high-dimensional data such as images, the computational complexity becomes daunting and the performance deteriorates rapidly. To reveal those relations, a new type of method based on deep learning for images clustering is needed. Thus we choose InfoGAN to be the clustering model in our framework.
%The main goal of generative modelling is to learn a generator distribution $P_{G}(x)$ that matches the real data distribution $P_{data}(x)$. Instead of trying to explicitly assign probability to every $x$ in the data distribution, GAN learns a generator network $G$, which generates samples from the generator distribution $P_{G}$ through transforming a noise variable $z\sim P_{noise}(z)$ into a sample $G(z)$. Besides an adversarial discriminator network $D$ is brought to distinguish between samples from the true data distribution $P_{data}$ and the generator`s distribution $P_{G}$ while the generator $G$ is trained by playing against $D$. Formally, the minimax game is given by the following expression:
%\begin{equation}
%\begin{split}
%        min_{G}max_{D}V(D,G)&={}\mathds{E}_{x\sim P_{data}}[logD(x)] {}\\ &+\mathds{E}_{z\sim P_{noise}}[log(1-D(G(z)))]
%\end{split}
%\end{equation}

%One significant problem of GAN is that generator process the noise in a highly coupled mode, which breaches the correspondence between each single dimensions of noise variable $z$ and semantic features of data. To solve this problem, Xi Chen \textit{et al}. put forward Information Maximizing Generative Adversarial Network(InfoGAN)\cite{NIPS2016_6399} assuming semantic features of data can be partially controlled by a series latent codes including continuous ones and categorical ones.
Generative Adviseral Network (GAN) is proposed by Goodfellow \textit{et al} in 2014[17]. It aims to learn a generator distribution $P
_{G}(x)$ that matches the real data distribution $P_{data}(x)$. However, One significant problem of GAN is that generator process the noise in a highly coupled mode, which breaches the correspondence between each single dimensions of noise variable z and semantic features of data. To solve this problem, Xi Chen \textit{et al}. put forward Information Maximizing Generative Adversarial Network (InfoGAN)[18].

%In InfoGAN the authors thus split the noise vector $z$ into two parts: source of incompressible noise $z^{`}$ and latent codes $c$ which will target the salient structured semantic features of the data distribution. In order to distinguish which elements in $x$ does a semantic feature of the data distribution correspond, a classifier $Q$ is also brought in the model. Moreover, because of the high degree of freedom of generative learning, we can easily find a solution satisfying $P(x|c)=P(x)$, a case in which latent codes $c$ are invalid. Therefore, an information-theoretic regularization is proposed: there should be high mutual information between latent codes $c$ and generator distribution $G(z,c)$. Thus the objective function of InfoGAN is:
In InfoGAN the authors thus split the noise vector $z$ into two parts: source of incompressible noise $z'$ and latent codes $c$ which will target the salient structured semantic features of the data distribution. By controlling the value of some dimension of $c$ we can change the generative images in a certain aspect, such as the size of eyes and sparsity of hairs in images of human faces. We assume that the conditional probability model is $P_{G}(x|c)$. In order to distinguish which elements in $x$ does a semantic feature of the data distribution correspond, a classifier $Q$ is also brought in the model. Moreover, because of the high degree of freedom of generative learning, we can easily find a solution satisfying $P_{G}(x|c)=P_{G}(x)$, a case in which latent codes $c$ are invalid. Therefore, an information theoretic regularization is proposed: there should be high mutual information between latent codes $c$ and generator $G(z,c)$. Thus the objective function of InfoGAN is[17][18]:
\begin{equation}
    \min_{G}\max_{D}V_{I}(D,G)=V(D,G)-\lambda I(c;G(z,c))
\end{equation}

Considering that directly maximize the mutual information is not computationally tractable, variational mutual information maximization is utilized in which a guide distribution $Q$ (classifier mentioned above) is introduced[17][18]:
\begin{equation}
    \min_{G,Q}\max_{D}V_{InfoGAN}(D,G,Q)=V(D,G)-\lambda L_{1}(G,Q)
\end{equation}

\subsection{Our Framework}
Architectures are one of the most important factors in city image according to Kevin Lynch[2]. In this subsection we introduce in details the method of the proposed framework whose main goal is to unsupervised classify the buildings in real street view. Since InfoGAN can deconstruct images to semantic components and cluster images according to their semantic distances, not like other clustering methods using the geometric distances, our framework is based on InfoGAN.
The diagrammatic sketch of our framework is showed in Figure 1.

%\paragraph{data cleaning}We first put the raw data which contains many images without buildings at all into the InfoGAN for pre-train. By analysing the correspondence between semantic features of the raw images and latent codes we find out the categorical code determining whether there are any architectures in the raw images, namely clustering images with and without architectures. After this step we obtain a set of images with at least one building, which are regarded as our source database $O$.

\paragraph{Architectural Segmentation}To extract buildings correctly we need to segment different semantic elements first. As we introduce above, FCAN is adopted in the consideration of the unsupervised property as well as the domain adaptation problem in semantic segmentation, which introduces the idea of appearance-level and representation-level invariance into unsupervised adaptation to enhance the performance of semantic segmentation. We adopt the pre-trained ResNet-50[19] architecture as the basic CNN and then train a generator and a discriminator with samples in source domain and target domain, by synchronizing the discriminator with the generator. In this way we can reduce the prediction error in the target domain, and obtain a higher accuracy of semantic segmentation in street view images as a result.

Mask processing is implemented after gaining feature maps of images in $\mathbf{O}$, which aims to eliminate semantic features belonging to non-architectures. By setting pixels of non-architectures in feature maps to zero we obtain the masks $\mathbf{M}$ of the original images. $\mathbf{M}$ contains the information about contours and perspective relations of street buildings. 

\paragraph{Unsupervised Classification}Finally images in $\mathbf{M}$ are put into InfoGAN for unsupervised learning, in which the auxiliary distribution $Q$ is parametrized as a neural network. An conditional distribution $Q(c|x)$ is generated after the image pass through the fully connected layer since $Q$ and $D$ share the same parameters. That is to say, not only the computational cost of $Q$ can be ignored but also the correspondences between categories of generative images and specific categorical code in $c$ are able to be concluded. Source database $O$ is regarded as the test set put into InfoGAN which is also trained by $O$, and $Q$ is regarded as a "classifier", which score the test images according to the categorical latent codes $c$. The results of $Q$ are the clustering categories of street buildings, which reveal the intrinsic information hidden in the true data distribution. Not only can we correspond them with those existing in urban planing fields but also explore potential new standards for classification.

\section{Experiments}
In this section we study the street view images in Nanjing, China, as a typical application case of our framework. Besides, two improving schemes (Interpolation and Raster Fairy) are presented to further dig categorical information under different themes. As a baseline model, K-means is used to be compared with our framework to see how the framework in this paper improves traditional clustering methods.

\begin{figure}[b]
    \centering
    \begin{subfigure}[b]{0.22\textwidth}
        \includegraphics[width=\textwidth]{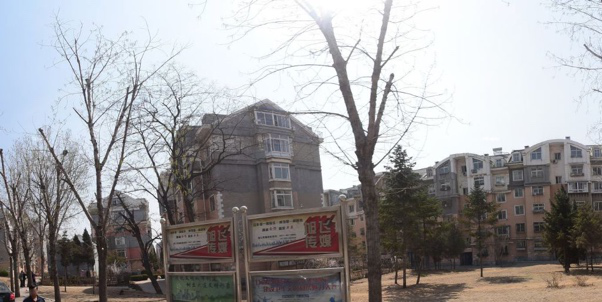}
        \caption{Origin}
        \label{fig:origin figure}
    \end{subfigure}
    \hfill
    \begin{subfigure}[b]{0.22\textwidth}
        \includegraphics[width=\textwidth]{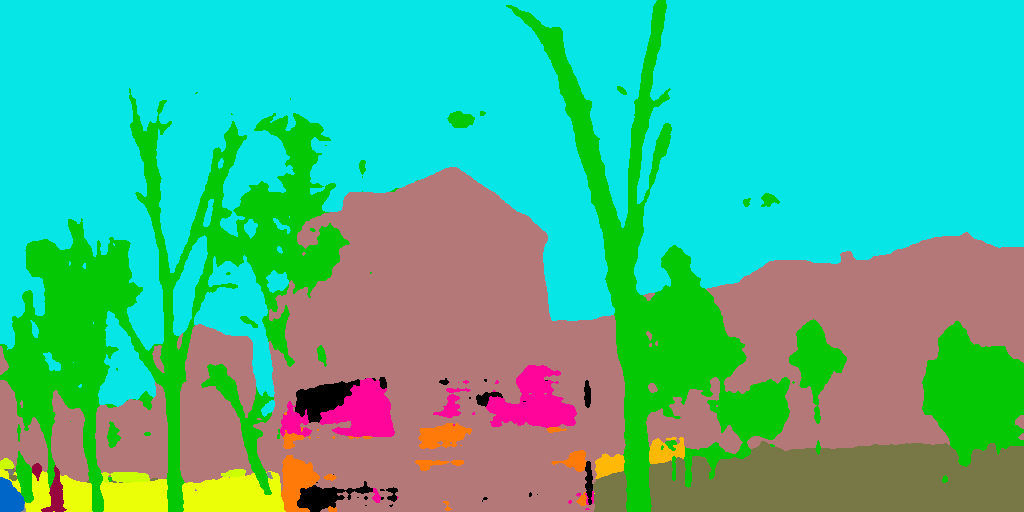}
        \caption{Segmentation}
        \label{fig:segmentation}
    \end{subfigure}
    
    \begin{subfigure}[b]{0.22\textwidth}
        \includegraphics[width=\textwidth]{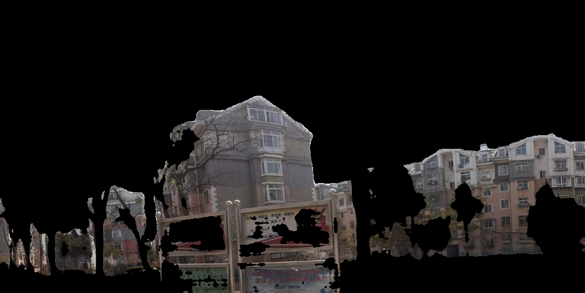}
        \caption{Mask}
        \label{fig:mask figure}
    \end{subfigure}
    \hfill
    \begin{subfigure}[b]{0.22\textwidth}
        \includegraphics[width=\textwidth]{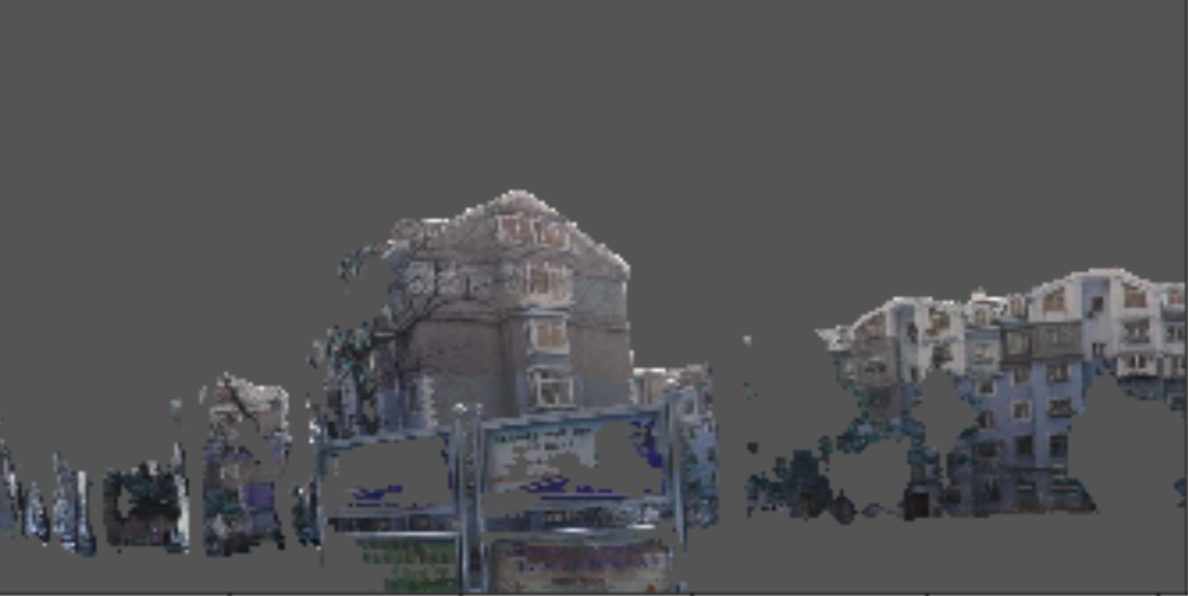}
        \caption{Interpolation}
        \label{fig:4-1-4}
    \end{subfigure}
    \caption{Middle results. (a) is the original figure; (b) is the result of semantic segmentation; (c) is the result of masking, (d)is the result of interpolation.}
\label{fig:4-1}
\end{figure}

\begin{figure}
    \centering
    \includegraphics[width=8cm]{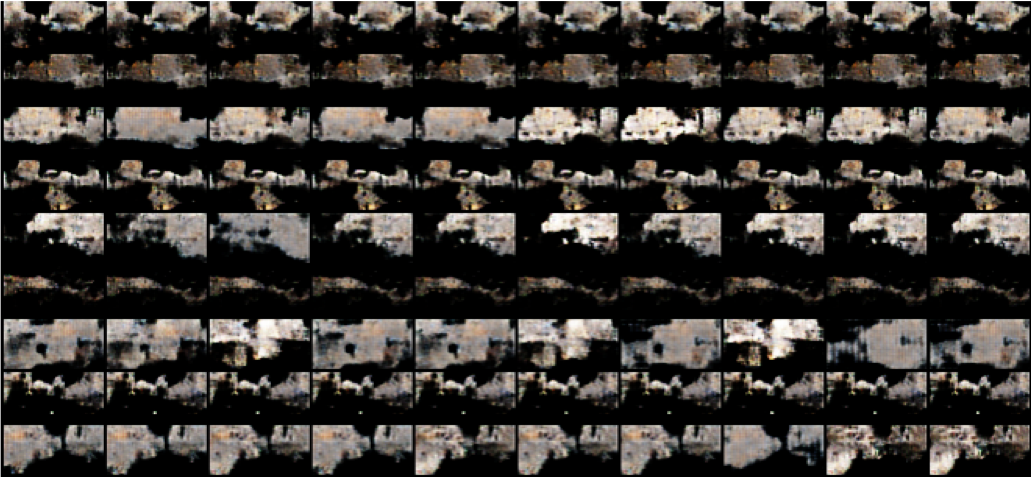}
    \caption{Generative Image. Each row represents a discrete category(only depict 9 in 25) and each column represents a continuous dimension of the input vector.}
    \label{fig:4-12}
\end{figure}

\subsection{Experimental Setting}
\paragraph{Hyperparameters}Our proposal is implemented on PyTorch1.0 framework and \textit{Adam}[20] is exploited to optimize the InfoGAN. The input images of discriminator $D$ and classifier $Q$ are color image of size $32\times 64$, while the input vector $z$ of generator $G$ includes three components: (i) categorical codes $c_{dis}$, whose size is uncertain; (ii) continuous codes $c_{con}$, with size $2\times 1$; (iii) incompressible noise $z_{rnd}$, with size $70\times1$. $Q$ and $D$ share all convolutional layers and there is one final fully connected layer to output parameters for the conditional distribution $Q(c|x)$. We use an up-convolutional architecture for the generator networks $G$                                                      . We use leaky rectified linear units (lRELU) as the nonlinearity applied to hidden layers of the discrminator networks, and normal rectified linear units (RELU) for the generator networks. Note that the loss function of the output layer in $D$ is binary cross entropy (BCE) loss for discrimination while the one in $Q$ is cross entropy loss in order to classification. We set initial learning rate to 2e-4 for $D$ and 2e-3 for $G$ and $Q$, and batchsize and epochs is set to 100 and 200 respectively. The tradeoff parameter $\lambda$ is set to 1. The leaky rate of lRELU is set to 0.1.

\begin{figure}[H]
    \centering
    \begin{subfigure}[b]{0.22\textwidth}
        \includegraphics[width=4cm,height=2.12cm]{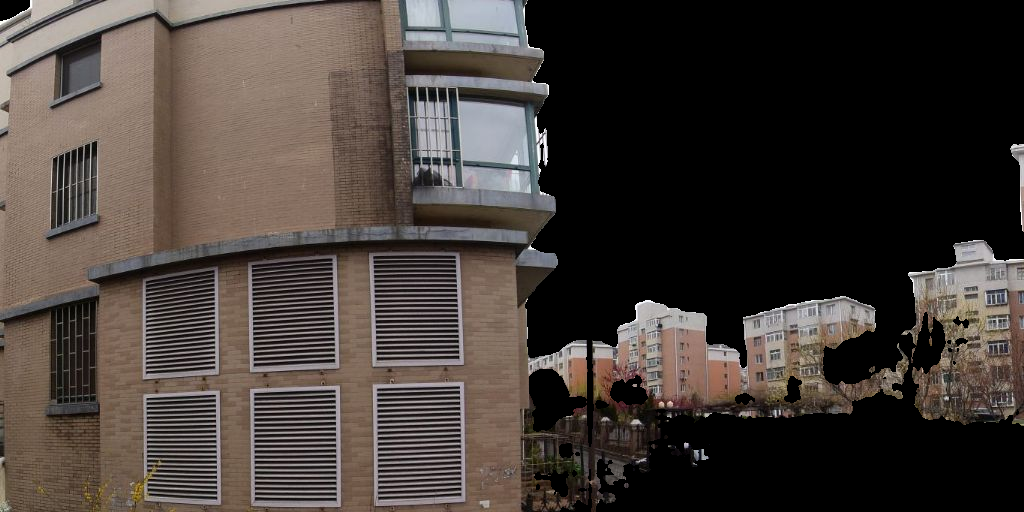}
        \caption{Original}
        \label{fig:origin figure}
    \end{subfigure}
    \hfill
    \begin{subfigure}[b]{0.22\textwidth}
        \includegraphics[width=\textwidth]{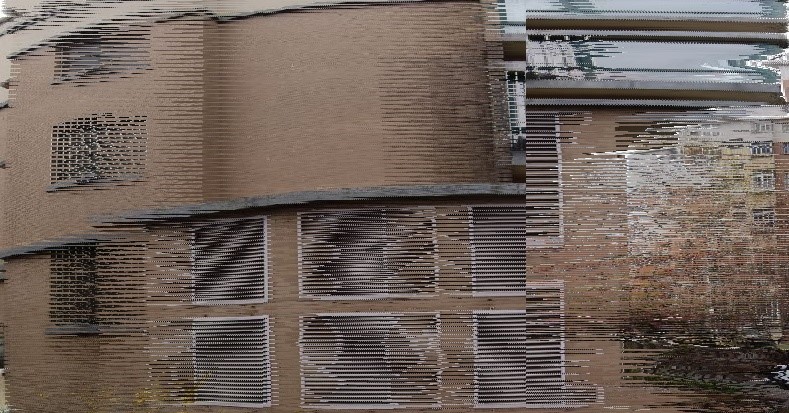}
        \caption{RF}
        \label{fig:segmentation}
    \end{subfigure}
\caption{Result of RF transformation.}
\label{fig:4-3}
\end{figure}

\begin{figure*}
    \centering
    \begin{subfigure}[b]{0.24\textwidth}
        \includegraphics[width=\textwidth,height=3.5cm]{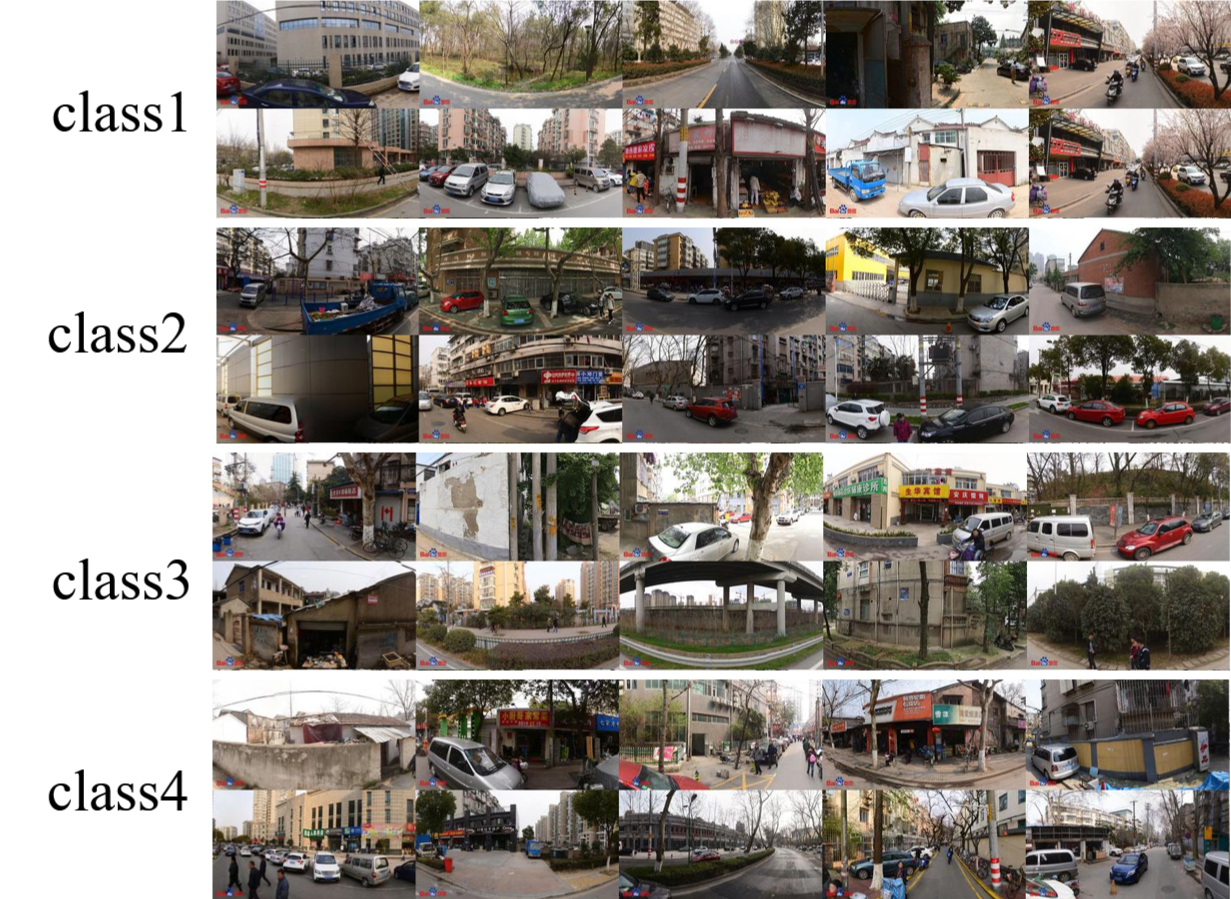}
        \caption{Kmeans}
        \label{fig:4-2-4}
    \end{subfigure}
    \hfill
    \begin{subfigure}[b]{0.24\textwidth}
        \includegraphics[width=\textwidth,height=3.5cm]{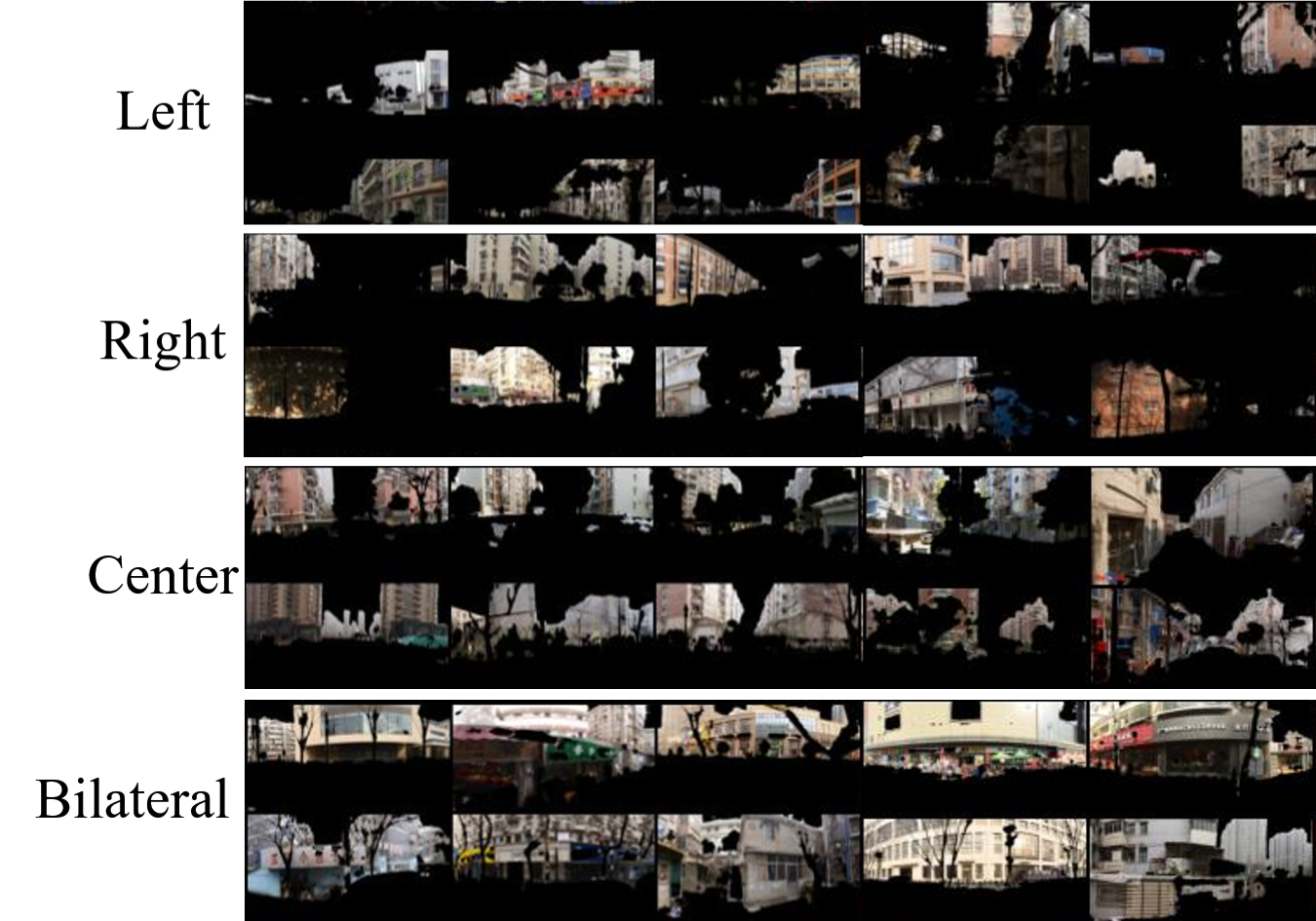}
        \caption{Perspectives}
        \label{fig:4-2-1}
    \end{subfigure}
    \hfill
    \begin{subfigure}[b]{0.24\textwidth}
        \includegraphics[width=\textwidth,height=3.5cm]{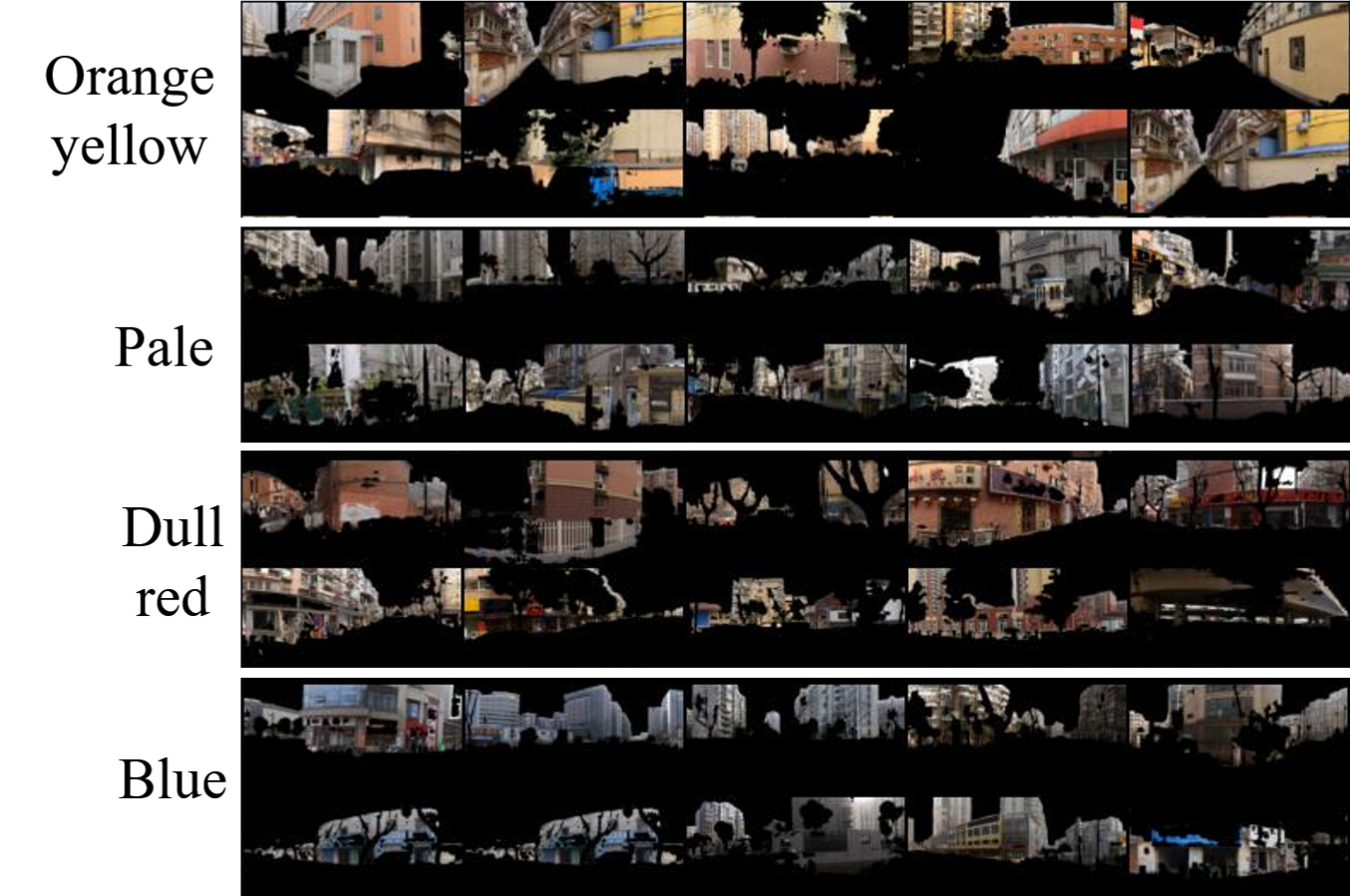}
        \caption{Color Systems}
        \label{fig:4-2-2}
    \end{subfigure}
    \hfill
    \begin{subfigure}[b]{0.24\textwidth}
        \includegraphics[width=\textwidth,height=3.5cm]{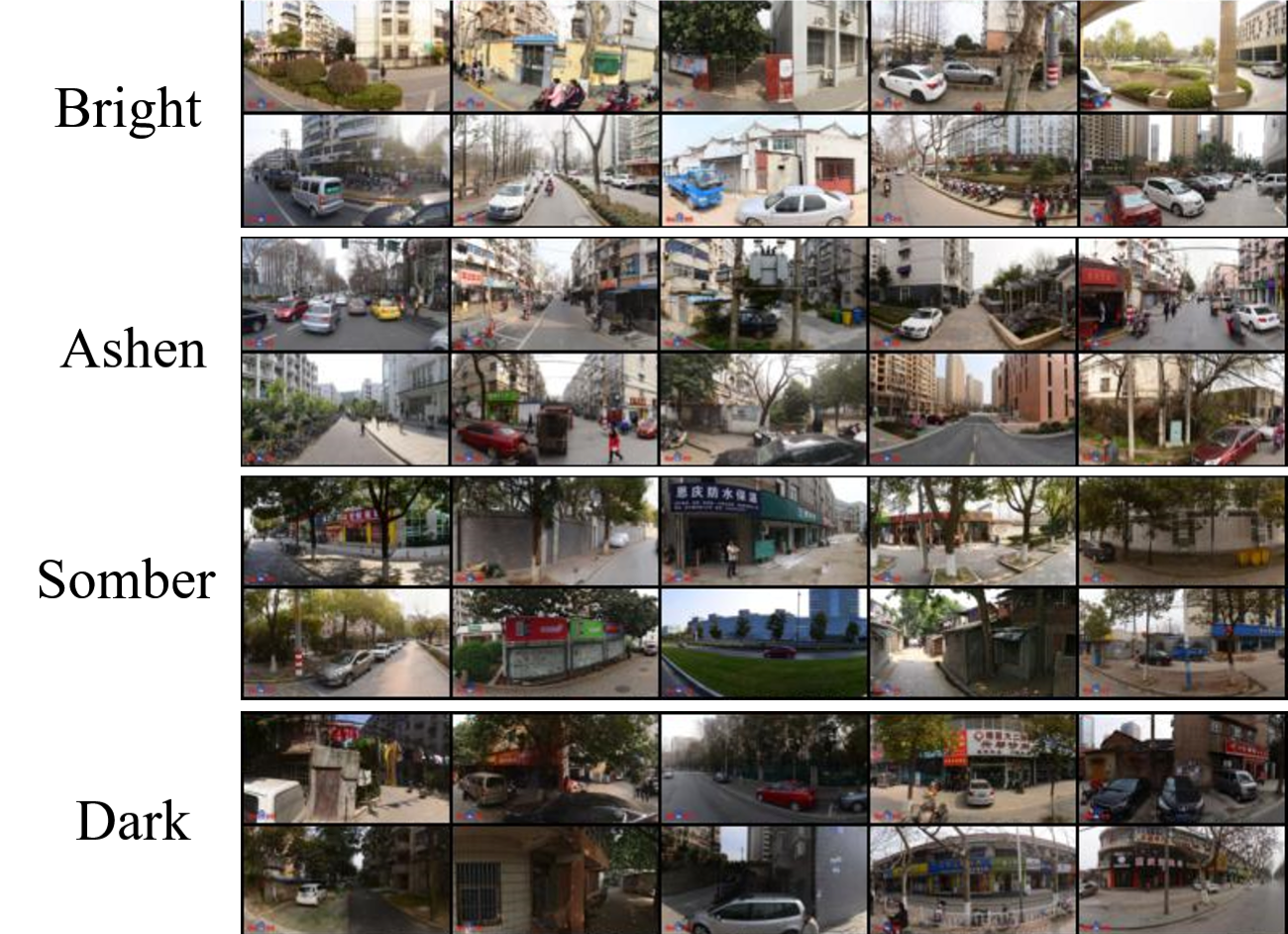}
        \caption{Gray Scales}
        \label{fig:4-2-3}
    \end{subfigure}
    \caption{Some clustering results. The architectures are clustered into several categories under four different themes: perspectives, gray scales, colors and functions. Different rows in (a) shows the classification results of Kmeans; Different rows in (b) shows different perspectives clustered from $\mathbf{M}$; Different rows in (c) shows different colors clustered from $\mathbf{R}$, the orange yellow buildings are mixed habitation buildings at the same time, similarly the pale buildings are high-rise residence buildings, the dull red buildings are urban business district buildings, and dark gray buildings are suburban commerce buildings; (d) shows different gray scales clustered from $\mathbf{I}$.}
\label{fig:4-2}
\end{figure*}
\paragraph{Datasets}The database we utilize consists of 372,606 street view images of Nanjing crawled from Baidu street view service. Three rounds of pre-training are carried out for eliminating images without buildings. The first round eliminate 294,951 images with a sheet of sky or greenland and the second round eliminates 59,522 images with meaningless streetscapes. After three rounds of cleaning, we get a source database $\mathbf{O}$ including 18,133 images with at least one building.

\paragraph{Trainning Strategy}In the processing of semantic segmentation, the FCAN is pre-trained by a small amount of labeled images in source domain and a large amount of unlabeled images in target domain. We employ four NVIDIA 1080TI GPUs to train the FCAN, with database of Cityscapes, which is a popular benchmark for urban street scenes containing high quality pixel-level annotations of 5,000 images collected in street scenes from 50 different cities. When we put the street view images of Nanjing into the trained FCAN, a typical result is shown as Figure 2 along with the mask. 

Images in $\mathbf{M}$ are put into InfoGAN with the size of categorical codes $c_{dis}$ set to 25, considering a smaller quantity of categories will lead to ambiguity in clustering results.When we finished the training process,we can get the generating graphs of every training epoch. Figure 3 is the generative image of the 100 epoch during training process with masks as inputs, in which each row represents a discrete category (only depict 9 in 25) and each column represents a continuous dimension of the input vector. From the image we can see that some information about colors and facades of architectures is gradually learned during the training process. Therefore, we put $\mathbf{O}$ into InfoGAN for test after training. As a result, we obtain 25 categories of images clustered by InfoGAN which are partly showed in Figure 5.

\subsection{Results and analysing}
%我们首先使用kmeans,对图片数据集M进行了聚类，得到如图Figure4(a)的聚类结果。研究发现，其聚类得到的每一类结果并不能展现出任何建筑风貌的特点，其轮廓系数计算得到为-0.024，从而可以发现Kmeans并不能对此进行有效的聚类。%
We first implement K-means to cluster image dataset $\mathbf{M}$, the result is shown in figure 5(a). From the result we can find that the categories clustered by K-means do not reveal any character of street architectures and the silhouette coefficient is calculated to be -0.024. This is to say that K-means is not a effect clustering method for our task.

Then we have preliminary clustering results of InfoGAN, which we will analyse and improve in this subsection. From the clustering results of $\mathbf{M}$ we can find it performances very well in clustering different perspectives of street buildings. However, when comes to other information like the facades, the results are completely chaos. In this case, we propose two improving schemes as follows:

% \noindent\items{\textit{Interpolation}: Replace the values of non-architectural pixels by the mean value of architectural pixels， in order to capture the brightness of architectural pixels. This processing is showed in Figure 2d

% \noindent\items{\textit{Raster Fairy}: To eliminate interference of architectural distribution, we compute the number of architectural pixels $N$ and evenly factorise $N$ into $N=W\times H$, $W$ and $H$ are used as size of the regularized images. Then bring them into RF transformation. After transformation the images are reshaped to a uniform size. This processing is showed in Figure 4
\textit{Interpolation}: Replace the values of non-architectural pixels by the mean value of architectural pixels, in order to capture the brightness of architectural pixels. This processing is showed in Figure 2d

\textit{Raster Fairy}: To eliminate interference of architectural distribution, we compute the number of architectural pixels $N$ and evenly factorise $N$ into $N=W\times H$, $W$ and $H$ are used as size of the regularized images. Then bring them into RF transformation. After transformation the images are reshaped to a uniform size. This processing is showed in Figure 4

The output images of Interpolation are denoted by $\mathbf{I}$ and the ones of RF transformation are denoted by $\mathbf{R}$. Table 1 shows the clustering results of $\mathbf{M}$ under the theme of perspectives. Similarly, Table 2 shows the clustering results of $\mathbf{I}$ under the theme of gray scales, while Table 3 shows the clustering results of $\mathbf{R}$ under the theme of color systems and architectural functions.
\begin{table}[htbp]
  \centering
  \caption{Clustering result of $\mathbf{M}$}
    \begin{tabular}{cc}
    \toprule
    perspectives & c\_dis \\
    \midrule
    left  & 0,1,2 \\
    right & 5,8,13,20,24 \\
    center & 3,6,11,15,16,17,19,22 \\
    bilateral & 4,7,9,10,12,14,18,21,23 \\
    \bottomrule
    \end{tabular}%
  \label{tab:4-1}%
\end{table}%

\paragraph{Perspective} From Table 1 we can find $\mathbf{M}$ being clustered into four categories (left, right, center, bilateral) according to the perspectives of the observer, which is also depicted in Figure 5. Since the factor of perspectives is regard as noise in the case of streetscape analysing or urban planning, classification of perspectives can be seen considered as a filter, which eliminates the interference brought by differences of perspectives.

\begin{table}[htbp]
  \centering
  \caption{Clustering result of $\mathbf{I}$}
    \begin{tabular}{cc}
    \toprule
    gray scale & \multicolumn{1}{c}{c\_dis} \\
    \midrule
    dark  & 2,6,7,9,10,14,15,17,19 \\
    somber & 0,8,12,13,16,21 \\
    ashen & 1,3,5,22,23 \\
    bright & 4,11,18,20 \\
    \bottomrule
    \end{tabular}%
  \label{tab:4-2}%
\end{table}%

\paragraph{Gray Scale} Table 2 shows that images in $\mathbf{I}$ are clustered into four categories revealing the gray scale of the image: bright, ashen, somber and dark, which is also depicted in Figure 5. As far as we can see, the difference of the gray scale of an architectural image has no effect on our goal. Therefore, we design the scheme of RF transformation to obtain more meaningful results.

\begin{table}[htbp]
  \centering
  \caption{Clustering result of $\mathbf{R}$, the meaning of abbreviations: HR, high-rise residence; LS, low-rise residence in suburbs; UH, Urban residence; LU, low-rise residence in urban areas; UB, urban business district; SC, suburban commerce; BD, business district; BN, bridge noise }
   \begin{tabular}{ccc}
    \toprule
    \multicolumn{2}{c}{RF} & \multicolumn{1}{c}{{c\_dis}} \\
    Color  & Function  & \multicolumn{1}{c}{} \\
    \midrule
    Pale  & HR    & 0,1,9,12,17 \\
    grayish yellow & LS    & 2,10,19,24 \\
    Orange yellow & UH    & 3,7,14 \\
    Bright white & LU    & \multicolumn{1}{c}{4} \\
    Dull red & UB    & 5,13 \\
    dark gray & SC    & 6,8,11,22,23 \\
    blue  & BD    & 15,21 \\
    Hybrid & BN    & 16,18,20 \\
    \bottomrule
    \end{tabular}%
  \label{tab:4-3}%
\end{table}%

\paragraph{Color System and Architectural Function} As Table 3 shows, results of RF transformation, $\mathbf{R}$, are clustered into eight color systems. Surprisingly, we find each color system correspond a certain architectural function respectively. We randomly select 200 buildings of different colors and their functions confirm our conclusion. That is to say, not only physical categories are clustered but also architectural functions are clustered by our framework, which are of guiding significance in the case of large-scale architectural classification. We depict this result in Figure 5. For instance, in Figure 5(c), the color of the building is orange yellow,  and most of the buildings are urban residential buildings. Besides, the pale embodies in high-rise residential buildings.
 %我们使用四种方法得到的聚类结果对图片打标签，然后将图片按照3:1的方式分为训练集和验证集，使用有监督学习的方式对图片进行训练和测试，实验测试结果如Table \ref{tab:4-4}所示：
%\paragraph{Effect Validation}Four clustering methods are used to label images, and then the pictures are divided into training set and verification set according to 3:1. The pictures are trained and tested by supervised learning. The experimental results are shown in Table \ref{tab:4-4}.
\paragraph{Effect Validation}Images in $\mathbf{M}$ are labeled by four clustering methods (K-means, Info-GAN (Mask), Info-GAN (Interpolation) and Info-GAN (RF)). We divide these labeled images $\mathbf{L}$ into training set and verification set according to 3:1, and then we use the training set to train a supervised learning model (ResNet-18)[19] and calculate the validation accuracy rates on verification set. We assume that a higher classification accuracy rate comes from a better clustered training set. These accuracy rates are shown in Table
The pictures are trained and tested by supervised learning. The experimental results are shown in Table 4.
\begin{table}[htbp]
  \centering
  \caption{Comparison of Clustering Methods}
    \begin{tabular}{ccccc}
    \toprule
    Method & K-means & Mask & Interpolation & RF \\
    \midrule
    val acc  & 0.560 & 0.804 & 0.826 & 0.825\\
    \bottomrule
    \end{tabular}%
  \label{tab:4-4}%
\end{table}%

From   these   results   we   can   see   that   the   architectures are   clustered   into   several   categories   under   four   different themes (perspectives,  gray  scales,  colors  and  functions),  the result images of each theme come from different preprocessings. From the clustering effect, we can find that Info-GAN is much better than K-means. Moreover, not only physical categories are clustered but also architectural functions are clustered by our framework, which are of guiding significance in the case of large-scale architectural classification.


\section{Conclusions}
In this paper, we present a practical framework to generate street architectural categories from existing street view images and then to score each image in all learned clusters, which for the first time applying InfoGAN on street architectural classification. By employing this framework, we overcome the defects of supervised methods to a large extent. We implement this framework in an experiment of streetscapes in Nanjing, China. As a result, the street architectures are clustered into several categories of four different themes(perspectives, gray scales, colors and functions). The clustering categories of each theme are easily recognized by eyes. 

Based on work in this paper, a more detailed clustering result can be obtained by implementing our framework iteratively on the existing outcomes. Meanwhile more preprocessing algorithms except interpolation and RF will be utilized in the future work. Furthermore, through experiments based on this framework, we will see if any new categories of street architectures can be learned by machine.

% \bibliographystyle{ieeetr}

% \bibliography{reference1}

\end{document}